\documentclass[letterpaper, 10 pt, conference]{ieeeconf}  

\IEEEoverridecommandlockouts                              

\title{\LARGE \bf
LS-HAR: Language Supervised Human Action Recognition with Salient Fusion, Construction Sites as a Use-Case
}

\author{Mohammad Mahdavian$^{1}$, Mohammad Loni$^{2}$, Ted Samuelsson$^{2}$, and Mo Chen$^{1}$
\thanks{$^{1}$Mohammad Mahdavian and Mo Chen are with Computer Science Department,
        Simon Fraser University, Burnaby, Canada, 
        \texttt{\small \{mmahdavi, mochen\}@cs.sfu.ca}}%
\thanks{$^{2}$Mohammad Loni and Ted Samuelsson are with Future Solutions Department, Volvo Construction Equipment, Eskilstuna, Sweden, \texttt{\small \{mohammad.loni, ted.samuelsson\}@volvo.com}%
}
}
\usepackage{textcomp}
\usepackage{gensymb}
\usepackage{url}
\usepackage{algorithm}
\usepackage{algpseudocode}
\usepackage{hyperref}

\usepackage{amssymb}
\usepackage{amsmath}
\usepackage{graphicx}
\usepackage{subcaption}
\usepackage{multirow}
\usepackage{array}
\usepackage{pifont}

\let\labelindent\relax
\usepackage{enumitem}
\usepackage{xspace}
\usepackage{verbatim}
\usepackage{wrapfig}
\usepackage{lipsum}
\usepackage{xcolor}
\usepackage{arydshln}

\newcommand{\cmark}{\ding{51}}
\newcommand{\xmark}{\ding{55}}

\definecolor{nmcolored}{rgb}{1.0,0.0,0.0}

\definecolor{mlcolored}{rgb}{0.0,0.0,0.0}

\definecolor{mmcolored}{rgb}{0.0,0.0,0.0}
\newcommand{\mm}[1]{\textcolor{mmcolored}{#1}}

\definecolor{mmiroscolored}{rgb}{0.0,0.0,0.0}
\newcommand{\mmi}[1]{\textcolor{mmiroscolored}{#1}}

\newlist{mylist}{enumerate*}{1}
\setlist[mylist]{label=(\roman*)}

\newcommand{\VCEdataset}{VolvoConstAct\xspace} 

\begin{document}

\maketitle
\thispagestyle{empty}
\pagestyle{empty}

\begin{abstract}

Detecting human actions is a crucial task for autonomous robots and vehicles, often requiring the integration of various data modalities for improved accuracy. In this study, we introduce a novel approach to \mmi{Human Action Recognition (HAR) using language supervision named LS-HAR} based on skeleton and visual cues. Our method leverages a language model to guide the feature extraction process in the skeleton encoder. Specifically, we employ learnable prompts for the language model conditioned on the skeleton modality to optimize feature representation. Furthermore, we propose a fusion mechanism that combines dual-modality features using a salient fusion module, incorporating attention and transformer mechanisms to address the modalities' high dimensionality. This fusion process prioritizes informative video frames and body joints, enhancing the recognition accuracy of human actions. Additionally, we introduce a new dataset tailored for real-world robotic applications in construction sites, featuring visual, skeleton, and depth data modalities, named \VCEdataset. This dataset serves to facilitate the training and evaluation of machine learning models to instruct autonomous construction machines for performing necessary tasks in real-world construction sites. To evaluate our approach, we conduct experiments on our dataset as well as three widely used public datasets: NTU-RGB+D, NTU-RGB+D 120, and NW-UCLA. Results reveal that our proposed method achieves promising performance across all datasets, demonstrating its robustness and potential for various applications. The code, dataset, and demonstration of real-machine experiments are available at: 
 
 \noindent \url{https://mmahdavian.github.io/ls_har/}  

\end{abstract}

\section{INTRODUCTION}
\label{sec:introduction}

    Autonomous machines and robots, a rapidly growing industry, must interact effectively with humans using visual cues, auditory commands, or manual controllers. For hazardous environments like construction sites, visual instruction signals are particularly effective \cite{wang2021vision}. Researchers have explored various methods for recognizing human actions, including body pose \cite{chen2021channel, chi2022infogcn, duan2022revisiting}, visual cues \cite{wang2021actionclip, nguyen2014stap, rao2022denseclip}, or a combination of both \cite{ahn2023star, bruce2021multimodal} \mm{using available public datasets~\cite{liu2019ntu,wang2014cross}}.
    Skeletons provide detailed limb movement information, while videos offer rich contextual cues such as objects. However, skeleton-based methods require additional deep learning techniques and struggle with actions involving object interactions, while video-based methods are susceptible to environmental disturbances like image blur. Combining these modalities addresses each other's deficiencies, enhancing overall performance. 

    Recent advancements in machine learning architectures have significantly improved HAR techniques. Researchers have employed cutting-edge frameworks such as transformers \cite{vaswani2017attention}, Vision Transformers (ViT) \cite{dosovitskiy2020image}, and Graph Convolutional Networks (GCN) \cite{kipf2016semi} to refine these methodologies. The GAP \cite{xiang2023generative} framework, for instance, enhances skeleton-based HAR by incorporating semantic information from textual content to supervise the skeleton modality. This method uses a contrastive loss, similar to that in the CLIP \cite{radford2021learning}, to minimize the discrepancy between the skeleton data and the corresponding text vector generated by a transformer-based \cite{vaswani2017attention} language model.

\begin{figure}[t] 
    \centering
    \includegraphics[width=0.95\columnwidth]{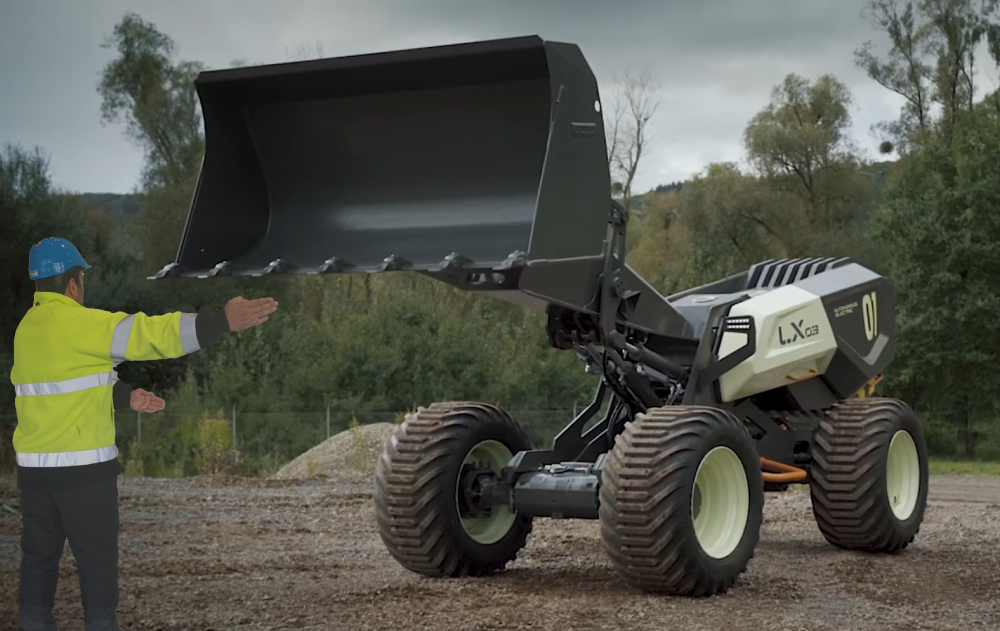} 
    \caption{A construction worker providing instructions to the Volvo autonomous wheel loader, demonstrating the integration of manual guidance and automated processes for higher operational efficiency and safety.}
    \label{fig:use-case_overview}\vspace{-5mm}
\end{figure}

    Similarly, ActionClip \cite{wang2021actionclip} applies this principle to image and text pairs, using a temporal transformer architecture \cite{lim2021temporal} integrated with CLIP to enhance HAR. In both GAP and ActionClip frameworks, the language model acts as a knowledge engine, aiding training and enriching the dataset's contextual understanding. \mm{However, effectively combining these two modalities under the supervision of language models can lead to significant improvements in the field of HAR.} In this paper, we introduce a novel end-to-end methodology that integrates human skeleton data and video for HAR, supervised by a language model for global skeleton and limb-based features. Our studies show that our visual modules perform well independently, whereas the skeleton modality relies more on text supervision. The CoOp framework \cite{zhou2022learning} highlights the impact of prompt selection on model performance, advocating for learnable prompt vectors combined with action labels instead of static prompts. CoCoOp \cite{zhou2022conditional} further refines this by conditioning the learnable prompt on the other modality. We adopt this strategy to enhance supervision on our skeleton encoder using dynamically adjusted prompts during training.

    Integrating features from diverse modalities is crucial for enhancing model accuracy. Traditionally, this has been done using summation \cite{bruce2022mmnet} or CNNs \cite{guo2023b2c}. However, recent advances show that transformers are highly effective for this purpose \cite{ahn2023star}.
    We propose a novel transformer-based fusion approach that prioritizes key features from different modalities. Our method incorporates a fine-grained down-sampling module to highlight valuable data segments and utilizes attention mechanisms to enhance their importance, leading to improved HAR performance. 
    
    \mm{In addition, our method for Human Activity Recognition (HAR) in construction sites addresses the growing interest in this area, driven by its potential to enhance productivity through the analysis of operations~\cite{han2013empirical,cheng2023construction} and to improve safety, such as by predicting worker fatigue~\cite{ibrahim2023investigating}. A significant challenge in this field is the limited availability of multi-modal datasets tailored for machine instruction in construction environments. Previous research has relied on datasets that are not specifically adapted to these settings. To overcome this limitation, we propose \VCEdataset, a new dataset designed for HAR in construction workers. This dataset aims to improve the robustness and generalization of robotic systems in real-world construction environments.}

    \textbf{Contributions}: \begin{mylist} \item We introduce a novel approach to dual-modality HAR, leveraging language models as knowledge engines. \item Employing learnable prompts conditioned on the skeleton features, we supervise and enhance global skeleton and limb feature extraction. \item Our innovative fusion technique implicitly identifies the frames and body joints most crucial for HAR, enhancing feature fusion. \item We present, for the first time, a dataset tailored for construction sites to recognize human actions for instructing machines. This dataset aims to enhance safety and efficiency on construction sites by providing data to monitor worker actions, predict hazards, and optimize workflows.  
    \end{mylist}

\section{Related Works}
\label{sec:relatedWorks}


\subsection{Machine Learning for HAR} \label{sec:relatedWorks: HAR}
    Human motion studies and HAR can generally be categorized into skeleton-based, video-based, and skeleton-video-based methods. \textbf{Skeleton-based} methods primarily use CNNs \cite{wang2021skeleton}, graph-based approaches \cite{peng2020learning, peng2021rethinking,anh2024enhanced}, or attention mechanisms \cite{song2017end, mahdavian2023stpotr, nikdel2023dmmgan} to model joint relationships and skeleton motion. CNN-based methods \cite{duan2022revisiting} employ 3D heat maps generated from 2D skeletons instead of GCNs to represent human body structure, followed by CNN layers for further processing. This approach demands significant memory, as each body joint requires a separate image layer.
    \textbf{Video-based} methods traditionally utilize CNNs as effective local feature learners \cite{simonyan2014two} or combine RGB images with optical flow streams \cite{carreira2017quo}. Notable methods in this domain include Action Machine \cite{zhu2018action}, $\pi$-ViT \cite{reilly2023just}, and DVANet \cite{siddiqui2024dvanet}.
    \textbf{Skeleton-Visual-based} methods leverage multiple modalities to enhance HAR performance \cite{davoodikakhki2020hierarchical, das2020vpn, bruce2022mmnet, ahn2023star}. MMNet \cite{bruce2022mmnet} integrates skeleton joints and bones with RGB images as spatio-temporal regions of interest, enhancing HAR by focusing on key image regions linked to specific joints, though it requires additional pre-processing. Also, STAR-Transformer \cite{ahn2023star} uses spatio-temporal cross-attention to integrate RGB images and skeleton data.

\subsection{Applications of Language Models in HAR} \label{sec:relatedWorks:ApplicationsHAR}

   Since the introduction of CLIP \cite{radford2021learning}, recent studies have focused on integrating text with other modalities to leverage semantic knowledge and enhance HAR by bridging modality gaps. In CLIP and related methods \cite{radford2021learning, wang2021actionclip, xiang2023generative}, a contrastive loss transfers semantic context from text to other modalities. ActionCLIP \cite{wang2021actionclip} extends CLIP by adding a temporal transformer for HAR, while GAP \cite{xiang2023generative} combines Graph Convolutions (GC) with a Multi-scale Temporal Convolution (MTC) network as an encoder, supervised by both ground truth (GT) and a language model-generated text vector.

    \mmi{On a related note, Xu et al.~\cite{xu2025language} leveraged an LLM assistant to train a skeleton-based model. It maps LLM knowledge into a priori global relationship (GPR) and a priori category relationship (CPR) topologies. GPR refines bone representations to highlight key node information, while CPR simulates category priors in the brain, using a PC-AC module for additional supervision to enhance class-distinguishable features.}
    

\subsection{Applications of HAR in Construction Sites} \label{sec:relatedWorks:construction}

To ensure safety and long-term health, visual HAR in hazardous industrial environments like construction sites has gained significant attention \cite{li2022action, wang2021vision, wang2023gaze, wu2023thermal, wang2024context}. Early studies focused on semantic relations between construction entities to enhance safety in human-robot collaborations \cite{kim2019semantic, kim2021toward}. Several studies used machine learning models to recognize workers' activities in construction site surveillance videos \cite{luo2018towards, yang2023transformer, li2022action, zhou2024construction, redmon2016you}. But, these systems are primarily for observation and security, with interactive machine control often limited to hand gestures \cite{wang2021vision, wang2023gaze, wu2023thermal, wang2024context}. Also, many studies did not publicly release their datasets, hindering reproducibility.
To address this, the Construction Motion Library (CML) \cite{tian2022construction} was released and is widely used for training HAR models in construction settings \cite{tian2024lightweight, yang2023transformer}. But, it has several limitations:
\begin{mylist}
\item Lacks visual data from real construction operations. \item Fails to capture the perception challenges of construction sites (e.g., poor visibility, sensor vibrations). \item Despite its diversity, it does not encompass all actions used to instruct construction machines.
\end{mylist}

\begin{figure*}
    \centering
\begin{subfigure}{0.6\textwidth}
    \centering
    \includegraphics[width=1\linewidth,trim={0cm  0cm 0cm 0cm },clip]{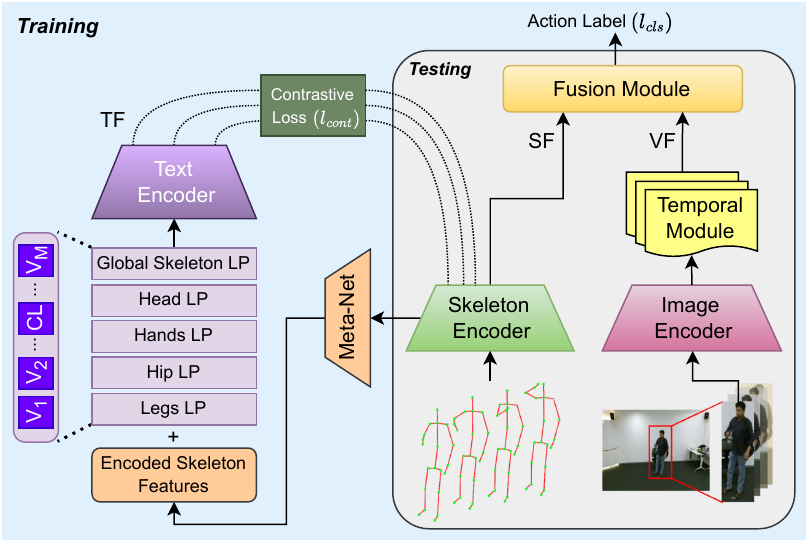}
    \caption{}
    \label{subfig:main_model}
    \vspace{-2mm}
\end{subfigure}
\hspace{1cm}
\begin{subfigure}{0.215\textwidth}
    \centering
    \includegraphics[width=1\linewidth,trim={0cm  0cm 0cm 0cm },clip]{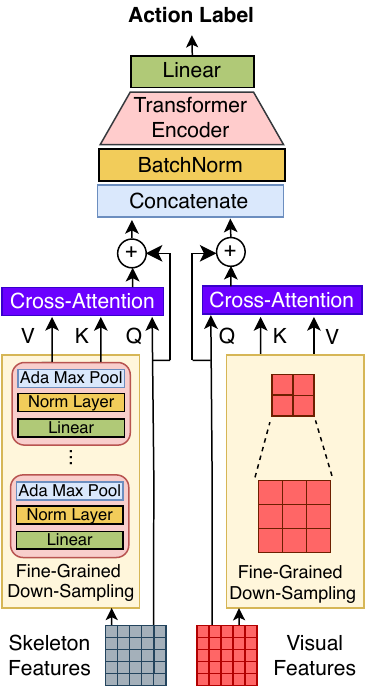}
    \caption{}
    \label{subfig:fusion_module}
    \vspace{-2mm}
\end{subfigure}
\caption{(a) Our model consists of text, skeleton, and image encoders, alongside a temporal visual module and a fusion module to integrate skeleton features (SF) and visual features (VF). During training, the text encoder uses learnable prompts (LP) to guide the skeleton encoder. A meta-net encodes skeleton features and adds them to the LPs (formatted as $V_1V_2...[CLASS]...V_M$, where each $V_M$ is a 512D vector) for different limbs (global, head, hands, hip, legs) to support text feature (TF) extraction. During testing, the text encoder is removed, and the model uses only RGB images and skeleton data.
(b) Our fusion module combines skeleton and visual features by emphasizing the most relevant aspects of each modality. The fine-grained down-sampling module is applied similarly to both, but skeleton features require two layers, while visual features only need one.}

\vspace{-5mm}
\label{subfig:main_figs}
\end{figure*}


\section{Methodology}

\label{sec:methodology}

In this section, we explain the technical details of our proposed model. As depicted in Fig.~\ref{subfig:main_figs}, we introduce a skeleton-video based HAR model, which integrates language models as a knowledge engine to supervise the skeleton encoder and the incorporation of learnable prompts (LP) facilitates enhanced utilization of the language models. Fig.~\ref{subfig:main_model} shows the main model structure. In the training phase, our model is fed with a sequence of RGB images ($V \in {\mathbb{R}}^ {T \times H \times W \times 3}$) capturing human actions, alongside the relative sequence of human 3D skeleton joints ($J \in {\mathbb{R}}^ {T \times j \times N \times 3}$). Here, $T$ represents the number of frames, $j$ the number of skeleton joints, and $N$ the number of humans. Also, $H$ and $W$ represent image height and width. Two separate encoders are employed to process data from the visual and skeleton modalities. Our GC-MTC-based skeleton encoder is responsible for learning the skeleton modality. Also, the visual encoder includes a pre-trained ViT-based image encoder and a transformer-based temporal module to learn the relation between the image features.

To further guide the learning process, a pre-trained transformer-based text encoder, such as BERT~\cite{devlin2018bert} or CLIP-text-encoder~\cite{radford2021learning}, is employed to supervise both global skeleton features and those specific to individual limbs. This is achieved using separate learnable prompt vectors: one for the overall skeletal structure and one for each distinct body limb section (head, hands, hips, and legs). To enhance the effectiveness of the text encoder's guidance, the learnable prompts are conditioned on the extracted skeleton features. This is accomplished by adding a feature vector derived from the skeleton data using the Meta-Net module with each learnable prompt vector.
    
To bridge the semantic gap between skeleton or limb features and text vectors, we use a contrastive loss function, $\mathcal{L}_\text{cont}$. Also, a fusion module combines the encoded skeleton and visual features, allowing for action prediction by calculating the loss against the GT, $\mathcal{L}_\text{cls}$, defined as:
 \vspace{-4mm}

     \begin{align}
        \mathcal{L}_\text{cls} = -y \log p_{\theta}(x)
     \end{align}

\vspace{-2mm}

     \noindent where $y$ is a one-hot vector of the ground-truth labels and $p_{\theta}(x)$ is the model's predicted probability distribution based on the fused features ($x$). Also, in order to calculate the $\mathcal{L}_\text{cont}$, it is necessary to first calculate the contrasting text-skeleton vector pairs for each picked batch of data in two directions.

 \vspace{-5mm}
 \begin{align}
    p_i^{s2t}(s_i) = \frac{\exp(\text{sim}(s_i,t_i)/\tau)}{\Sigma^{Batch}_{j=1} \exp(\text{sim}(s_i,t_j)/\tau)}  \\
    p_i^{t2s}(t_i) = \frac{\exp(\text{sim}(t_i,s_i)/\tau)}{\Sigma^{Batch}_{j=1} \exp(\text{sim}(t_i,s_j)/\tau)}     
 \end{align}
 
 \noindent where $s$ and $t$ represent extracted skeleton and text features. Also, sim function is a cosine similarity matrix defined between the two features, and $\tau$ denotes the temperature matrix. The $\mathcal{L}_\text{cont}$ is then calculated as follows:

\vspace{-3mm}
 \begin{align}
    \mathcal{L}_\text{cont} = \frac{1}{2}\mathrm{E}_{s,t\sim D}[KL(p^{s2t}(s),y^{s2t})+KL(p^{t2s}(t),y^{t2s})]
 \end{align}
 \noindent where $y^{s2t}$ and $y^{t2s}$ are similarity scores calculated based on GT, assigning a probability of 1 for positive pairs and 0 for negative pairs. To better capture the skeleton features of different limbs, we apply the same $\mathcal{L}_\text{cont}$ between the limb skeleton features and the corresponding text-encoded features. We then average these to obtain the total contrastive loss, $\mathcal{L}_\text{Tcont}$.
 The total training loss is defined as
 
\vspace{-5mm}

  \begin{align}
    \mathcal{L}_\text{total} = \lambda \mathcal{L}_\text{Tcont} + \mathcal{L}_\text{cls}
 \end{align}

 \noindent where $\lambda$ is a pre-defined constant variable (here assigned as 0.8).
    In the testing phase, the text encoder is removed from the model, leaving only the skeleton-visual modalities and the fusion module to perform the HAR task. In general, any model structures can be used to encode skeletons, images, or text, as the focus of this paper is on the overall concept. However, we provide detailed explanations of our specific model structures.
    
\vspace{-2mm}

\subsection{Skeleton Encoder}
    Our skeleton encoder combines GC and MTC components. The GC framework within facilitates the discernment of spatial relationships among body joints. Meanwhile, the MTC component is dedicated to capturing the temporal dynamics inherent within the sequence of skeletons.

\textbf{Graph Convolutions (GC).}
    The human skeleton is considered as a graph, $G=({V,\epsilon})$, where $V$ represents body joints, and $\epsilon$ signifies the edges connecting them. At a given layer $l$, the features of human joints, characterized by a feature dimension of $F$, are expressed as $H^l \in {\mathbb{R}}^{j \times F}$. The GC operation can be defined as follows:
    
  \vspace{-4mm}

    \begin{align}
    H^{l+1} = \sigma(D^{-\frac{1}{2}} A D^{\frac{1}{2}} H^{l} W^l)
    \end{align}

    Here, $\sigma$ represents the activation function, $D \in {\mathbb{R}}^{j \times j}$ is the degree matrix, $A \in {\mathbb{R}}^{j \times j}$ denotes the adjacency matrix, and $W^l$ encapsulates the learnable parameters of the $l$-th layer.

\textbf{Multi-scale Temporal Convolutions (MTC).}
    To capture the temporal aspect of human skeletal movements, we use the MTC \cite{chen2021channel, liu2020disentangling}, which has four branches: a MaxPool branch, two branches with different dilation rates (1 and 2), and a branch with only 1 $\times$ 1 convolutions. Each branch integrates a 1 $\times$ 1 convolution to reduce channel dimensionality, and their outputs are combined to produce the final result.

    \vspace{0mm}
\subsection{Visual Encoder}

    To encode RGB images and extract relevant features, we use the CLIP-video-encoder \cite{radford2021learning} based on ViT \cite{dosovitskiy2020image}, which employs attention mechanisms to highlight visual context and detect global features, including scene objects. The ViT image encoder divides images, represented as $I \in {\mathbb{R}}^{H \times W \times 3}$, into fixed-size patches $I_p \in {\mathbb{R}}^{N_p \times (P^2 \times 3)}$, with $P$ as the patch size and $N_p$ as the total number of patches. The patches undergo linear embedding and are then processed by a transformer model with positional embedding. We test two variants: ViT-B/32 and ViT-B/16. ViT-B/32 divides images into 32 $\times$ 32 patches, while ViT-B/16 uses 16 $\times$ 16 patches, allowing finer detail detection at the cost of longer training.
    
\textbf{Video Temporal Module.}
    Our image encoder extracts RGB image features, and to capture the temporal dynamics of the image sequence, we use a temporal module based on ActionClip. This module, similar to the main transformer structure \cite{vaswani2017attention}, integrates positional embeddings with the image features and processes them through layers of temporal transformers to generate video modality features.

\vspace{0mm}

\subsection{Text Encoder}

In our study, we use a text encoder based on the CLIP-text-encoder \cite{radford2021learning} or BERT \cite{devlin2018bert} to encode input prompts related to the global skeleton and each body limb. These prompts are tokenized and processed through transformer blocks to generate feature vectors representing the textual descriptions. During end-to-end training, the text encoder adapts to produce text vectors from the prompts, which may be manually crafted or generated by a Large Language Model (LLM) \cite{xiang2023generative}. We employ learnable prompts conditioned on the skeleton modality to enhance model adaptation for HAR.


\textbf{Learnable Prompts.}
    To optimize text encoder performance, we use learnable vectors for prompts rather than pre-defined ones. As shown in Fig.~\ref{subfig:main_model}, the prompts are formatted as $V_1V_2...[CLASS]...V_M$, where each $V_M$ (with $m \in {1, \ldots, M}$) is a 512-dimensional vector (for CLIP), initially set to ``X''. The $[CLASS]$ label can appear at the beginning, middle, or end of the prompt, and $M$ represents the number of learnable tokens. To enhance supervision of skeleton features, we use distinct learnable prompts for each body section. A Meta-Net encoder, $h_\theta(.)$, parameterized by $\theta$, incorporates a vector derived from the skeleton features into these prompts by adding them together. The Meta-Net consists of a linear layer, a ReLU activation, and another linear layer. Each context token is then obtained by $V_M(s) = V_M + \pi$, where $s$ denotes skeleton features and $\pi = h_\theta(s)$. This approach improves guidance for the text encoder.

    Notably, conventional approaches train learnable prompts while freezing the text and visual encoders, focusing solely on learning the prompt vector. Our method departs from this practice, enabling one-shot, end-to-end training.

\subsection{Fusion Module}

    To leverage the strengths of both skeleton and video modalities, we design a fusion module, shown in Fig.~\ref{subfig:fusion_module}, to utilize the unique advantages of each. Simple summation or concatenation would only average the performance of the two modalities. Therefore, a method is needed to dynamically prioritize the most salient features from each modality.

    To effectively fuse the two modalities, it is essential to extract the most relevant information from each due to the substantial overlap and high dimensionality of the features. We address this with a fine-grained down-sampling module, which uses linear transformation, normalization, and adaptive max pooling to iteratively select the most informative features. In each iteration, adaptive max pooling retains only the most significant half of the features, reducing dimensionality. This module is initially applied independently to both skeleton and video modality features. For skeleton features, which initially consist of 100 features each of dimension 256, two iterations of pooling reduce the count from 100 to 25. For visual features, one iteration reduces the count from 16 to 8, yielding the best results.

    After dimensionality reduction, we apply a cross-attention mechanism to each modality. This mechanism evaluates the importance of each feature relative to the down-sampled representation, determining the significance of each feature within its modality. The calculated weights are then added to the corresponding features. The feature vectors from both modalities are concatenated and passed through a transformer encoder after a batch normalization layer. The resulting tensor is averaged along the feature dimension and fed into a fully-connected layer to reduce the dimension to the number of class labels.



\section{Action Recognition for Instructing Autonomous Construction Machines}


Heavy machinery, such as wheel loaders and excavators, plays a crucial role in construction tasks like digging, lifting, and material handling. These machines demand skilled operators, which can be labor-intensive and costly. Construction sites are challenging environments, characterized by high noise levels, extreme temperatures, and dynamic conditions \cite{hamid2008causes}, complicating machine operation and increasing error risks. With advancements in robotics, there is growing interest in developing autonomous construction machines to improve safety and efficiency. Traditionally, manual machines receive instructions from workers positioned outside the machine (Fig.~\ref{fig:use-case_overview}). For instance, extending both hands and bending the elbows signals the machine to move forward. To enhance safety and task efficiency, autonomous construction machines need to understand and act on worker instructions, necessitating the integration of a HAR model.

HAR in construction sites encounters challenges due to poor visibility and adverse weather conditions, which complicate the accurate interpretation of workers' actions. Safety protocols also require workers to maintain a safe distance from machines, making it difficult to discern small-sized objects. These factors, combined with the lack of a well-defined dataset for construction site communication, limit the effectiveness of existing HAR techniques trained on public datasets. To advance HAR in construction contexts, we introduce \VCEdataset, a new public dataset specifically designed for HAR in construction sites. \VCEdataset stands out from existing HAR datasets by addressing construction-specific challenges, defining a new set of actions, and encompassing a diverse range of construction environments.


\section{Experiments}
\label{sec:experiments}

\subsection{Datasets}
\label{sec:experiments:dataset}
    \vspace{-1mm}

\textbf{\VCEdataset} dataset (ours) comprises RGB images, depth data, and 2D and 3D human skeletons. These skeletons are extracted using the human detection capabilities of the ZED 2i camera. The dataset features 17 distinct actions performed by construction workers instructing a vehicle. Motions are captured in 20 unique videos containing specific details, and every recorded motion is divided into 5 sections. We use 75\% of the total recorded videos from each action as the training set and the remaining 25\% as the test set. Additional details about the dataset are provided in Table~\ref{tab:VCE_Usecase}.  

\begin{table}[t] 
    \centering
     \caption{Specifications of the \VCEdataset dataset.}
    \label{tab:VCE_Usecase}\vspace{-2mm}
    \begin{tabular}{m{2.7cm}|m{5cm}} \hline
        {\textbf{Criteria}} & {\textbf{Specification}} \\ \hline
        
         {Actions / Instructions} & Emergency, Stop, Slow Down, Come, Back Up, Turn Left, Turn Right, Do You See Me?, All OK?, Follow Me, Arm Circle, Excavator Swing Left, Excavator Swing Right, Load Lift Up, Load Lift Down, Wheel Loader Tilt Up, Wheel Loader Tilt Down \\ \hline

         Camera Model & Stereo RGB+D ZED 2i Camera \footnotemark\\ \hline
         Resolution & 1280 $\times$ 720 \\ \hline
         Frame Rate & 30 FPS \\ \hline
         
         Samples per Action & 20 \\ \hline
         Sample Duration & 10-14 Sec \\ \hline
         Number of 2D \& 3D skeleton joints  &  34\\ \hline
         Subjects & 2 \\ \hline
         
    \end{tabular}\vspace{-5mm}
\end{table}

\footnotetext{\url{https://www.stereolabs.com/en-se/store/products/zed-2i}}

\textbf{NTU-RGB+D} \cite{shahroudy2016ntu} is a public dataset widely used for video and/or skeleton based human motion studies and contains 56,880 skeletal action sequences for 60 action classes. The benchmarks for evaluation include Cross-Subject (XSub) and Cross-View (XView). In the XSub evaluation, both training and testing sets comprise data sourced from 20 distinct subjects, ensuring generalizability of the models trained on this dataset. Conversely, the XView evaluation leverages samples captured by cameras 2 and 3, totaling 37,920 sequences for training purposes. The remaining 18,860 sequences, captured by camera 1, are reserved for the test set in this evaluation framework.

\textbf{NTU-RGB+D 120} \cite{liu2019ntu} extends the NTU-RGB+D dataset with an additional 57,367 video and skeleton sequences, capturing an extra 60 action classes, resulting in a total of 120 action classes. Two benchmark evaluations exist for this dataset, including Cross-Subject (XSub) and Cross-Setup (XSet) settings.

\mm{\textbf{NW-UCLA} \cite{wang2014cross} is a publicly available small dataset consisting of 1,484 action sequences performed by ten actors across ten different classes. The skeleton data contains 20 joints and 19 bone connections and was captured using three Kinect v1 cameras. Following the standard benchmark outlined in \cite{wang2014cross}, we used data from cameras 1 and 2 for training, while data from camera 3 was reserved for testing.} 

\vspace{-1mm}

\subsection{Implementation Details}
\vspace{-1mm}
\label{sec:experiments:implementation}

In our experiments, using entire images as the visual modality did not yield optimal performance. However, applying a cropping technique to focus solely on the human subject significantly improved model performance and reduced the data volume. This method is particularly effective with ViT as the visual encoder, enabling the attention mechanism to focus on relevant objects related to human actions, such as a glass for the ``drinking water'' action. We used YOLOv5~\cite{redmon2016you} for human detection and cropping. For actions involving multiple subjects, all detected humans and the intervening area were included. During training and testing, all image frames were resized to uniform dimensions.
For skeleton data from the NTU-RGB+D and NTU-RGB+D120 datasets, we applied the denoising techniques described in \cite{zhang2020semantics, chi2022infogcn}. For the NW-UCLA dataset, we followed the standard procedure outlined in \cite{xiang2023generative}. Also, we used 16 frames for both RGB images and skeleton modalities, uniformly selecting data from the total frames of the video and the associated skeleton frames, as detailed in \cite{ahn2023star} and \cite{wang2018temporal}.

In our paper, we employ and evaluate both ViT-B/16 and ViT-B/32 architectures from CLIP's visual encoder as the backbone for our visual encoder. For our learnable prompts vector, we set the number of context tokens M to 24. Our fusion module incorporates attention layers and transformer blocks with 8 heads and a dimension of 512.
We train our model on a single A40 GPU with a batch size of 16 with AdamW optimizer and over 55 epochs \mm{for all datasets except NW-UCLA, where the model was trained for 20 epochs}. The learning rate is set to $5 \times 10^{-6}$ for pre-trained parameters (text and visual encoders) and $5 \times 10^{-5}$ for all other learnable parameters. A learning rate warm-up is applied for the first 5 epochs, followed by a decrease by a factor of 10 at the 35th and 45th epochs.

\vspace{-1mm}

\subsection{Results} \label{sec:experiments:results}
\vspace{-1mm}

In Table~\ref{NTU}, we present a comparative analysis of our method against several state-of-the-art (SOTA) approaches on the NTU-RGB+D, NTU-RGB+D120 and NW-UCLA datasets. For a fair comparison, we primarily focus and restrict our observation in Table~\ref{NTU} on methods that do not require pre-training of their modality encoders. As depicted in Table~\ref{NTU}, we achieve better or comparable performance with respect to the SOTA methods. \mmi{Specifically, for NTU60-XSub, we improve model performance by 1\% compared to the closest similar SOTA method~\cite{bruce2021multimodal} and 0.7\% compared to the best pose-based method~\cite{chi2022infogcn}. Also, for the NTU60-XView, NTU120-XSub and UCLA benchmarks, our results and the best SOTA are comparable.} 

\mmi{It is important to highlight that we fine-tuned our model and hyperparameters specifically for the NTU60-XSub benchmark, which explains the small performance gap between our results and the best SOTA results on other benchmarks. The performance gap between our method and the best SOTA is more pronounced on the NTU120-XSet benchmark. This discrepancy can be attributed to differences in camera setups and view angles between our training and testing datasets. Our approach involves cropping the human from each image frame, which can potentially reduce the similarity between the training and testing samples, thereby impacting model performance. Therefore, this does not indicate a lower overall quality of our presented method compared to the SOTA.} 

\mmi{Additionally, methods such as TSMF~\cite{bruce2021multimodal}, STAR~\cite{ahn2023star}, and STAR++~\cite{ahn2023star++} use heatmaps to learn 2D skeletons, requiring significantly more GPU memory during inference than our skeleton encoder. In contrast, our approach employs GCNs to efficiently process skeleton data, making it more suitable for embedded systems in construction machinery, where GPU memory is often constrained in edge computing devices.}

\mmi{We also trained our model using the \VCEdataset dataset and compared its performance with some of the SOTA methods that have publicly available open-source code suitable for embedded systems in construction machines. We achieved better results compared to the SOTA methods as shown in Table~\ref{volvo_results}. It is important to mention that our goal is to establish a baseline for action recognition of construction workers using our dataset. Enhanced performance for the skeleton modality on \VCEdataset could be achieved with additional denoising and pre-processing, as we used raw skeleton data to test the model’s robustness and establish our evaluations as a baseline. Additionally, our high performance on video only accuracy suggests that future studies could benefit from focusing solely on the visual modality, which is more advantageous for long-range HAR.}

\begin{figure}[b]
\vspace{-5mm}
    \centering
    \includegraphics[width=0.98\linewidth]{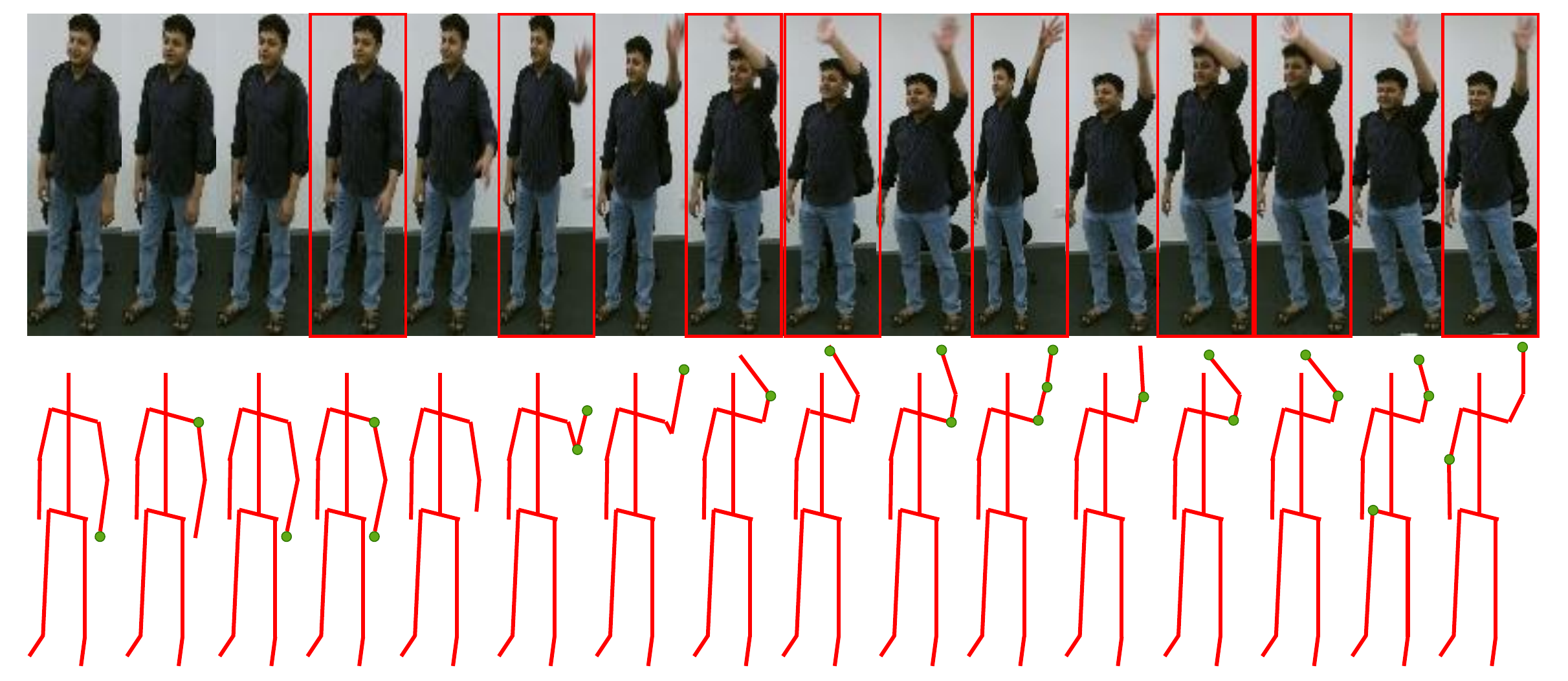}\vspace{-2mm}
    \caption{A sample of the image frames and skeleton joints picked by the fine-grained down-sampling module showing the most informative data for detecting the human action.}
    \label{fig:image_skeleton}\vspace{0mm}
\end{figure}

Also, Fig.~\ref{fig:image_skeleton} presents a qualitative result showing selected images and skeleton joints for a sequence depicting a human performing a ``waving'' action. The most effective visual features are associated with frames exhibiting significant hand movements, while the skeleton features related to the waving arm have the greatest impact on action detection.

\begin{table} [t]

  \centering
   \caption{Model performance comparison with SOTA on the NTU-RGB+D~\cite{shahroudy2016ntu} (NTU60), NTU-RGB+D120~\cite{liu2019ntu} (NTU120) and NW-UCLA (UCLA)~\cite{wang2014cross} datasets. $R$ and $P$ show the RGB and pose modalities and $PT$ specifies pre-trained models.}\vspace{-2mm}
\begin{tabular}{p{1.7cm}p{1.2cm}p{0.4cm}p{0.6cm}p{0.4cm}p{0.4cm}p{0.5cm} }

 \hline
\multirow{2}{*}{\textbf{Method}}  & \textbf{Modality} & \multicolumn{2}{c}{\textbf{NTU60 (\%)}} & \multicolumn{2}{c}{\textbf{NTU120 (\%)}} & \textbf{UCLA}\\
 \cline{3-6}
  &  \textbf{(PT)}  & {XSub } & {XView } & {XSub} & {XSet} & \textbf{(\%)} \\
\hline
 PoseC3D~\cite{duan2022revisiting} & P (\cmark)  & 93.7 & 96.6 & 86.0 & 89.6 & -\\
 VPN~\cite{das2020vpn} & R+P (\cmark)  & 95.5 & 98.0 & 86.3 & 87.8 & 93.5\\
 MMNet~\cite{bruce2022mmnet} & R+P (\cmark)  & 96.0 & 98.8 & 92.9 & 94.4 & 93.7\\
 VPN++~\cite{das2021vpn++} & R+P (\cmark)  & 96.6 & 99.1 & 90.7 & 92.5 & 93.5\\
 PoseC3D~\cite{duan2022revisiting} & R+P (\cmark)  & 97.0 & 99.6 & 95.3 & 96.4 & -\\
 \hline
MMST~\cite{cheng2023multi} & P (\xmark) & 89.6 & 95.3 & 85.3 & 86.0 & 95.3\\
HAR-ViT~\cite{han2024human} & P (\xmark) & 91.0 & 96.7 & 87.6 & 89.0 & -\\
CTR-GCN~\cite{chen2021channel} & P (\xmark) & 92.4 & 96.8 & 88.8 & 89.3 & 96.5\\
Info-GCN~\cite{chi2022infogcn} & P (\xmark) & 93.0 & 97.1 & 89.8 & 91.2 & 97.0\\
DC-GCN~\cite{bruce2021multimodal} & R+P (\xmark) & 90.8 & 96.6 & 86.5 & 88.1 & 95.3 \\
STAR~\cite{ahn2023star} & R+P (\xmark) & 92.0 & 96.5 & 90.3 & 92.7 & -\\
TSMF~\cite{bruce2021multimodal} & R+P (\xmark) & 92.5 & 97.4 & 87.0 & 89.1 & -\\
STAR++~\cite{ahn2023star++} & R+P (\xmark) & 92.7 & 97.6 & 90.9 & 92.9 & -\\
\hline
\textbf{Ours} & R+P (\xmark) & 93.7 & 97.3 & 90.3 & \mm{89.4} & 96.9\\
\hline
\end{tabular}
\label{NTU}
\vspace{-2mm}
\end{table}

\vspace{-1mm}

\begin{table} [t]
\centering
\caption{\mmi{Model performance comparison with SOTA on \VCEdataset (Volvo) dataset.}}\vspace{-2mm}
\begin{tabular}{p{3.6cm} >{\centering\arraybackslash}p{2.4cm}} 
 \hline
{\textbf{Method}} & {\textbf{\VCEdataset (\%)}} \\ 
\hline
VideoMAE v2 small~\cite{wang2023videomae} & 70.5  \\
VideoMAE v2~\cite{wang2023videomae} & 86.2  \\
VideoSwin~\cite{liu2022video} & 96.0  \\
PoseC3D~\cite{duan2022revisiting} & 94.8 \\
\hline
\textbf{Ours} (Video Only) &  92.9 \\
\textbf{Ours} with ViT-B/32 & 91.5\\
\textbf{Ours} & 96.4 \\
\hline
\end{tabular}\vspace{-5mm}
\label{volvo_results}
\end{table}


\subsection{Ablation Studies} \label{sec:experiments:ablations}

In this section, we examine the performance of each module using the NTU-RGB+D dataset, summarized in Table~\ref{ablations}. We first evaluate the efficacy of individual modalities (skeleton or visual) for HAR, both with and without a text encoder. Our analysis shows that the visual modality performs better, while the text encoder enhances the performance of the skeleton encoder.
Notably, merging the two modalities with our fusion module shows that using text supervision for both or neither modality decreases performance by 0.5\% and 0.6\% on XSub benchmarks compared to using it only for the skeleton encoder. This is due to the superior performance of our visual module and the fusion module's effectiveness, making text supervision unnecessary for the visual encoder.

In another aspect of our study, we assessed our fusion module without the salient feature extraction, relying instead on self-attention for each modality's features. Our ablation study revealed that extracting salient features improves model performance by 0.8\%. Additionally, learnable prompts enhance performance by 0.3\%, and conditioning on skeleton features adds another 0.1\%, \mmi{both of which remain valuable given our competitive benchmark}. For comparison, using the ViT-B/32 image encoder resulted in a 2.3\% performance decrease.


\begin{table} [t]
\centering
\caption{Ablation study for our method on the NTU-RGB+D~\cite{shahroudy2016ntu} (NTU60) dataset.}\vspace{-2mm}
\begin{tabular}{p{6.3cm}p{0.3cm}p{0.3cm}} 
 \hline
\multirow{2}{*}{\textbf{Method}} & \multicolumn{2}{c}{\textbf{NTU60 (\%)}}\\ \cline{2-3}
  &  {XSub} & {XView}\\
\hline
Only Skeletons & 86.2 & 92.5\\
\mm{Only Skeletons Supervised with Text} & 86.6 & 94.5 \\
Only Video & 91.2 & 96.5 \\
\mm{Only Video Supervised with Text} & 91.3 & 96.0  \\
Fusion without fine-grained down-sampling & 92.9 & 97.0  \\
Ours without Text Supervision on Skeletons or Video & 93.1 & 96.7\\
Ours with \mm{Supervising both Skeletons and Video} & 93.2 & 96.5 \\
Ours with Non-learnable Prompts (Pre-defined Text Prompts) & 93.4 & 97.1 \\
Ours without Conditional Prompts \mm{(Without Adding Extracted Skeleton Features to the Learnable Prompts)} & 93.6 & 97.4 \\ \hline
\textbf{Ours} with ViT-B/32 & 91.4 & 95.7 \\
\textbf{Ours} & 93.7 & 97.3\\
\hline
\end{tabular}
\label{ablations}
\vspace{-4mm}
\end{table}

\vspace{-3mm}

\section{Conclusion}
\label{sec:conclusion}

This paper presents a novel methodology for HAR that leverages both skeleton and visual data. Our approach enhances HAR performance by integrating a text encoder to guide the skeleton encoder, using learnable prompts conditioned on extracted skeleton features. We also introduced a fusion module that effectively combines visual and skeleton features by focusing on the most salient information from each modality. Additionally, we introduced the \VCEdataset, a new HAR dataset tailored for autonomous robots in construction environments. We anticipate this dataset will be a valuable resource for researchers working on worker action recognition in construction sites.

\textbf{Limitation.} For the \VCEdataset dataset, data collection was limited to short and mid-range distances due to the ZED 2i camera's limitations in long-range human pose detection. Additionally, no language supervision was applied to the visual modality, as its performance was already strong. These areas are highlighted for future investigation.

\section*{ACKNOWLEDGMENT}

We sincerely acknowledge Mr. Mikael Fries for his contributions to real machine development and his insightful discussions.

\bibliographystyle{IEEEtran}
\bibliography{root}

\end{document}